\documentclass{article}

\usepackage{microtype}
\usepackage{graphicx}

\usepackage{subcaption}

\usepackage{tabularx}   
\usepackage{booktabs} 
\usepackage{multirow}
\usepackage[most]{tcolorbox}
\usepackage{listings}
\usepackage{longtable}

\usepackage{hyperref}

\usepackage[accepted]{icml2025}

\usepackage{amsmath}
\usepackage{amssymb}
\usepackage{mathtools}
\usepackage{amsthm}

\usepackage[capitalize,noabbrev]{cleveref}

\theoremstyle{plain}

\theoremstyle{definition}

\theoremstyle{remark}

\newcolumntype{L}[1]{>{\raggedright\arraybackslash}p{#1}}  
\newcolumntype{C}[1]{>{\centering\arraybackslash}p{#1}}    

\usepackage[textsize=tiny]{todonotes}

\icmltitlerunning{TabReason: A Reinforcement Learning-Enhanced Reasoning LLM for Explainable Tabular Data Prediction}

\begin{document}

\twocolumn[
\icmltitle{TabReason: A Reinforcement Learning-Enhanced Reasoning LLM for Explainable Tabular Data Prediction}



\icmlsetsymbol{equal}{*}

\begin{icmlauthorlist}
\icmlauthor{Tommy Xu}{equal,borealis}
\icmlauthor{Zhitian Zhang}{equal,borealis}
\icmlauthor{Xiangyu Sun}{equal,borealis}
\icmlauthor{Lauren Kelly Zung}{equal,borealis}
\icmlauthor{Hossein Hajimirsadeghi}{equal,borealis}
\icmlauthor{Greg Mori}{borealis}
\end{icmlauthorlist}

\icmlaffiliation{borealis}{RBC Borealis, Canada}

\icmlcorrespondingauthor{Hossein Hajimirsadeghi}{hossein.hajimirsadeghi@borealisai.com}
\icmlcorrespondingauthor{Tommy Xu}{tommy.xu@borealisai.com}

\icmlkeywords{Machine Learning, ICML}

\vskip 0.3in
]



\printAffiliationsAndNotice{\icmlEqualContribution} 

\begin{abstract}
Predictive modeling on tabular data is the cornerstone of many real-world applications. Although gradient boosting machines and some recent deep models achieve strong performance on tabular data, they often lack interpretability. On the other hand, large language models (LLMs) have demonstrated powerful capabilities to generate human-like reasoning and explanations, but remain under-performed for tabular data prediction. In this paper, we propose a new approach that leverages reasoning-based LLMs, trained using reinforcement learning, to perform more accurate and explainable predictions on tabular data. Our method introduces custom reward functions that guide the model not only toward better prediction accuracy but also toward human-understandable reasons for its predictions. The proposed method is evaluated on financial benchmark datasets and compared against established LLMs.
\end{abstract}

\section{Introduction}
\label{sec:intro}

Tabular data, organized in rows and columns, is fundamental across various domains such as finance. Predictive modeling from such data is a core machine learning task, traditionally led by models like gradient boosting machines~\citep{chen2016xgboost, prokhorenkova2018catboost} and neural networks~\citep{arik2021tabnet, hollmann2022tabpfn}
These models have been effective but often lack transparency and interpretability, which are critical in high-stakes applications where understanding the reasoning behind predictions is essential for trust, regulatory compliance, and decision-making. For instance, in financial risk assessment, explaining why a loan application was rejected can be crucial for customer satisfaction and legal requirements.

Large Language Models (LLMs) have transformed natural language processing with their ability to understand, generate, and reason about text in a human-like manner. Their capacity to explain thought processes makes them promising for enhancing both accuracy and explainability in prediction tasks. However, applying LLMs to tabular data, which is structured and numerical, presents challenges, as LLMs are primarily trained on unstructured text data. Recent research has started to bridge the gap between language models and structured data by applying large language models (LLMs) to tasks involving the prediction and understanding of tabular data~\cite{feng2023empowering, yin2023finpt, hegselmann2023tabllm, bordt2024elephants, Yang_Wang_Li_Sen_Li_Liu_2025}. However, existing approaches have primarily focused on improving prediction accuracy, with little or no emphasis on generating explanations. Moreover, these methods typically rely on either pre-trained LLMs, conventional fine-tuning or few-shot prompting.

Inspired by the recent success of reinforcement learning in DeepSeek models~\cite{shao2024deepseekmath, guo2025deepseek}, we introduce a novel framework that unifies tabular data prediction with natural language explanations, optimized through reinforcement learning. The proposed method aims to achieve state-of-the-art accuracy while providing explainability through the model's reasoning process. By training the LLM with reinforcement learning, where the reward function is based on both prediction accuracy and the output quality, the approach seeks to create a model that excels in both performance and interpretability. 

This is particularly relevant for applications in financial assessment where explainability can enhance trust and decision-making. Recently, there has been a growing interest in using LLMs in financial problems with tabular data~\citep{feng2023empowering, zhang2023tablellama, xie2024finben, xie2024open, yang2024unleashing}. However, none of the previous works have addressed reasoning and explainability when performing prediction tasks on the tabular data. 
The key contributions of our work can be summarized as follows.
\vspace{-1.4em}
\begin{itemize}
    \item \textbf{Explainable tabular prediction:} We introduce an LLM-based model for tabular data prediction that inherently provides explainability. The reasoning steps generated by our model offer an interpretable view into the decision-making process. To the best of our knowledge, this is the first integration of RL and reasoning LLMs for tabular data prediction, paving the way for future research built upon this approach.
    \item \textbf{Performance Benchmarking:} We evaluate our proposed method on financial benchmark datasets, including credit risk assessment, fraud detection, financial distress identification, and claim analysis. The results demonstrate that a relatively small RL-trained reasoning LLM has the potential to outperform well-established LLMs. However, further experimental studies are needed to draw more definitive conclusions.
\end{itemize}

\section{Proposed Method}
\label{sec:method}
This section presents our proposed method for tabular data prediction using LLMs. The core idea is to prompt an LLM to infer the target value based on the provided input attributes. To train the model, we employ reinforcement learning, where the objective is to maximize a reward function that captures both prediction accuracy and the quality of the model's responses. Specifically, we utilize the Group Relative Policy Optimization (GRPO) method~\citep{shao2024deepseekmath}, which is described in detail in the following section.

\subsection{Group Relative Policy Optimization (GRPO)}

GRPO is a reinforcement learning algorithm designed to improve the reasoning capabilities of large language models (LLMs) while reducing computational overhead. Unlike traditional methods such as Proximal Policy Optimization (PPO) that require a separate value (critic) network to compute the advantage function, GRPO uses \emph{group-based reward normalization} to compute a relative advantage. In essence, for a given input (or prompt) the model generates a \emph{group} of outputs, and the algorithm uses the statistics (mean and standard deviation) of the rewards within this group to standardize (or normalize) the reward signal.

\subsubsection{Intuition Behind GRPO}
For each prompt $q$, assume we sample a group of $G$ outputs
\[
\mathcal{O} = \{ o_1, o_2, \dots, o_G \}
\]
using the old policy $\pi_{\theta_\text{old}}(o|q)$. Each output $o_i$ is assigned a reward $R_i$ (for example, 1 for a correct answer and 0 for an incorrect one). The group statistics are computed as:
\[
\mu = \frac{1}{G}\sum_{i=1}^G R_i, \quad \sigma = \sqrt{\frac{1}{G}\sum_{i=1}^G (R_i - \mu)^2}\,.
\]
Then, the \emph{relative advantage} for each output is defined as:
\[
\hat{A}_i = \frac{R_i - \mu}{\sigma}\,.
\]
This normalized advantage reflects how much better (or worse) an output is compared to the average performance in the group.

\subsubsection{GRPO Objective Function}
GRPO updates the policy by optimizing a surrogate objective similar to PPO but computed over the group of outputs. For a generated output $o_i$ with tokens $\{o_{i,1}, o_{i,2}, \dots, o_{i,T_i}\}$, the \emph{per-token probability ratio} is given by
\[
r_{i,t}(\theta) = \frac{\pi_\theta(o_{i,t} \mid q, o_{i,<t})}{\pi_{\theta_\text{old}}(o_{i,t} \mid q, o_{i,<t})}\,.
\]
Then, the GRPO objective can be written as
\begin{align}
J_{\text{GRPO}}(\theta) &= \mathbb{E}_{q \sim P(Q),\, \{o_i\} \sim \pi_{\theta_{\text{old}}}(o|q)} \Bigg[
\frac{1}{G} \sum_{i=1}^{G} \sum_{t=1}^{T_i} \nonumber \\
&\quad \min \Big( r_{i,t}(\theta) \hat{A}_i, \text{clip}(r_{i,t}(\theta), 1{-}\epsilon, 1{+}\epsilon)\, \hat{A}_i \Big) \nonumber \\
&\quad - \beta\, D_{\text{KL}}\big(\pi_\theta(\cdot|q,o_i) \,\|\, \pi_{\text{ref}}(\cdot|q,o_i) \big) \Bigg]\,.
\label{eq:grpo}
\end{align}

\vspace{-0.8em}
\begin{itemize}
  \setlength\itemsep{0.0em}
  \item $\pi_\theta$ is the current policy with parameters $\theta$, and $\pi_{\theta_\text{old}}$ is the policy before the update.
  \item $\hat{A}_i$ is the normalized (relative) advantage for output $o_i$.
  \item The inner $\min$ and $\text{clip}(\cdot)$ operations serve to prevent the probability ratio from deviating too far from 1, thus ensuring a stable update.
  \item $D_{\text{KL}}(\cdot \,\|\, \cdot)$ is the Kullback-Leibler divergence penalty, and $\beta$ is a hyperparameter controlling its strength. The reference policy $\pi_{\text{ref}}$ (often set to the initial supervised fine-tuned model) prevents the new policy from drifting too far from a desirable baseline.
\end{itemize}

In summary, GRPO optimizes the per-token policy by:
\vspace{-0.8em}
\begin{enumerate}
  \setlength\itemsep{0.0em}
  \item Sampling multiple outputs for each prompt.
  \item Computing the group mean and standard deviation of the rewards to obtain a relative advantage $\hat{A}_i$.
  \item Updating the policy using a PPO-like objective, with clipping and KL-penalty, but without requiring an explicit value function.
\end{enumerate}

This approach reduces the memory and computation requirements while effectively amplifying the probability of generating high-quality outputs, which is especially beneficial for large language models.

\subsection{Reward Modeling}
To optimize the model via reinforcement learning, we need to define reward functions. We use the following three types of rewards:
\vspace{-0.8em}
\begin{itemize}
    \setlength\itemsep{-0.3em}
    \item Response Format Rewards: This set of rewards evaluates if the model response follows the requested format. A positive reward (\texttt{0.5} in our experiments) is provided when the explanation is between \texttt{<reasoning>} and \texttt{</reasoning>} and the final prediction is between \texttt{<answer>} and \texttt{</answer>}.
    \item Answer Validity Reward: This reward evaluates whether the generated answer  matches \textit{one of} the expected answers. In our experiments, we use \texttt{0.5} as the reward value.
    \item Answer Correctness Reward: This reward evaluates if the final prediction (extracted from the expected answer format) is correct or not. The reward for correctness is set to \texttt{1.0} in our experiments. 

\end{itemize}

Note that the proposed framework is generic and supports the definition of arbitrary custom reward functions and values. For example, it is also possible to define rewards using other LLMs (as critics) or other ML models (as evaluators).

\section{Experiments}
\label{sec:experiments}
We conduct experiments on the financial assessment tasks introduced in~\cite{xie2024finben}, which provide a comprehensive benchmark for evaluating LLMs.

\subsection{Tasks and Datasets}
Table~\ref{tab:financial-datasets} provides an overview of the financial datasets used across different tasks, including credit scoring, fraud detection, financial distress identification, and claim analysis. Each dataset is listed with the number of test, train, and raw samples, along with the number of features available. The datasets vary significantly in size and complexity, ranging from small-scale datasets like German Credit and Australia to relatively larger datasets such as Lending Club and ccFraud. This diversity allows for comprehensive evaluation of models under varying data regimes and problem settings.
\begin{table}[htb]
\caption{Summary of datasets by task.}
\centering
\small
\setlength{\tabcolsep}{4pt}
\begin{tabular}{llcc}
\toprule
\textbf{Task} & \textbf{Dataset} & \textbf{Test/Train/Val} & \textbf{\#Features} \\
\midrule
\multirow{3}{*}{\shortstack[l]{Credit\\Scoring}} 
  & German        & 200/700/100           & 20 \\
  & Australia     & 139/482/69             & 14 \\
  & Lending Club  & 2691/9417/1345           & 21 \\
\midrule
\multirow{2}{*}{\shortstack[l]{Fraud\\Detection}} 
  & Credit Card Fraud   & 2279/7974/1139        & 29 \\
  & ccFraud       & 2098/7339/1048        & 7  \\
\midrule
\multirow{2}{*}{\shortstack[l]{Financial\\Distress}} 
  & Polish        & 1737/6076/868            & 64 \\
  & Taiwan Economic  & 1365/4773/681        & 95 \\
\midrule
\multirow{2}{*}{\shortstack[l]{Claim\\Analysis}} 
  & PortoSeguro   & 2382/8332/1190           & 57 \\
  & Travel Insurance   & 2534/8865/1266         & 9 \\
\bottomrule
\end{tabular}
\label{tab:financial-datasets}
\end{table}

\subsection{Prompting Templates}
We use one single system prompt for all the tasks as shown below.

\begin{tcolorbox}[colback=blue!5, colframe=blue!75!black, title=System Prompt, sharp corners, breakable, label={box:system_prompt}]
You are an expert in financial assessment.

Your task is to do assessment based on the financial status provided by attributes.

Respond in the following XML format with \texttt{<reasoning>} and \texttt{<answer>} tags:

\texttt{<reasoning>} \\
\texttt{...} \\
\texttt{</reasoning>} \\
\texttt{<answer>} \\
\texttt{...} \\
\texttt{</answer>}
\end{tcolorbox}

But, for each task, there is a separate query prompt customized for the target task as shown in Table~\ref{tab:query_prompts}.
\begin{table*}[h!]
\caption{Query prompts used for different financial datasets}
\small
\centering
\begin{tabular}{|l|p{13cm}|}
\hline
\textbf{Dataset} & \textbf{User Prompt} \\
\hline
German & ``Assess the creditworthiness of the following client as either 'good' or 'bad' based on the provided attributes." \\
\hline
Australian & ``Assess the creditworthiness of the following client as either 'good' or 'bad' based on the provided attributes. All the table attribute names including 8 categorical attributes and 6 numerical attributes and values have been changed to meaningless symbols to protect confidentiality of the data." \\
\hline
LendingClub & ``Assess the client's loan status as either 'good' or 'bad' based on the following loan records from Lending Club." \\
\hline
ccf & ``Detect the credit card fraud as either 'yes' or 'no' using the following financial table attributes. The attributes contains 28 numerical input variables V1, V2, …, and V28 which are the result of a PCA transformation and 1 input variable 'Amount' which has not been transformed with PCA. The feature 'Amount' is the transaction Amount, this feature can be used for example-dependent cost-sensitive learning." \\
\hline
ccfraud & ``Detect the credit card fraud as either 'yes' or 'no' using the following financial table attributes." \\
\hline
polish & ``Predict whether the company will face bankruptcy as either 'yes' or 'no' based on the following financial attributes." \\
\hline
taiwan & ``Predict whether the company will face bankruptcy as either 'yes' or 'no' based on the following financial attributes." \\
\hline
portoseguro & ``Determine whether to file a claim for the auto insurance policyholder as either 'yes' or 'no' based on the following table attributes of their financial profile. The table attributes that belong to similar groupings are tagged as such in the feature names (e.g., ind, reg, car, calc). In addition, feature names include the postfix bin to indicate binary features and cat to indicate categorical features. Features without these designations are either continuous or ordinal. Values of -1 indicate that the feature was missing from the observation." \\
\hline
travelinsurance & ``Determine the claim status as either 'yes' or 'no' based on the following table attributes for travel insurance status. The table attributes including 5 categorical attributes and 4 numerical attributes are as follows: Agency: Name of agency (categorical). Agency Type: Type of travel insurance agencies (categorical). Distribution Channel: Distribution channel of travel insurance agencies (categorical). Product Name: Name of the travel insurance products (categorical). Duration: Duration of travel (numerical). Destination: Destination of travel (categorical). Net Sales: Amount of sales of travel insurance policies (numerical). Commission: Commission received for travel insurance agency (numerical). Age: Age of insured (numerical)." \\
\hline
\end{tabular}
\label{tab:query_prompts}
\end{table*}

For textual representation of input attributes (features) in the query prompt, we follow the same format used in the FinBen benchmarks~\cite{xie2024finben}.

\subsection{Results}
\label{sec:results}
Table~\ref{tab:llm-performance} presents a comparative analysis of various large language models (LLMs) across multiple financial datasets using weighted F1 score
as evaluation metric. We trained TabReason using Qwen2.5-1.5B-Instruct~\citep{yang2024qwen2} as the base model, which is also included as a baseline in the table to demonstrate that RL-based tuning consistently enhances model accuracy across all datasets. The scores for other LLMs have been extracted from FinBen results~\cite{xie2024finben}. Overall, TabReason achieves the highest weighted F1 score on 7 out of 9 datasets. Among the other LLMs, no single model stands out as a clear winner. However, because most of these datasets are highly imbalanced, the weighted F1 score may not fully capture model performance. Financial tasks are typically highly imbalanced, presenting significant challenges for LLMs during both fine-tuning and evaluation. However, our model achieves particularly strong results on the LendingClub and Australian datasets, which are more balanced.
For a discussion of potential pitfalls, please refer to Appendix~\ref{sec:challenges}. For further information on LLM settings and RL tuning plots, see Appendix~\ref{sec:rl-tuning}.

Additionally, to demonstrate the explainability capabilities of TabReason, we show examples of its generated reasoning (i.e., explanations) and predictions in Appendix~\ref{sec:generation-examples} (Table~\ref{tab:generation-examples}).

\begin{table*}[!htb]
\caption{Performance comparison of various LLMs across different financial datasets using weighted F1 metric.}
\label{tab:llm-performance}
\centering
\scriptsize
\setlength{\tabcolsep}{2.2pt}
\begin{tabularx}{0.98\textwidth}{L{46pt} C{20pt} C{20pt} *{9}{C{26pt}} C{35pt} C{26pt} C{24pt}}
\toprule
\textbf{Dataset} & \textbf{Chat-GPT} & \textbf{GPT-4} & \textbf{Gemini} & \textbf{Llama2-7B-chat} & \textbf{Llama2-70B} & \textbf{Llama3-8B} & \textbf{FinMA-7B} & \textbf{FinGPT-7B-lora} & \textbf{InternLM-7B} & \textbf{Falcon-7B} & \textbf{Mixtral-7B} & \textbf{CFGPT-sft-7B-Full} & \textbf{Qwen2.5-1.5B} & \textbf{TabReason} \\
\midrule
German & 0.20 & 0.55 & 0.52 & \textbf{0.57} & 0.17 & 0.56 & 0.17 & 0.52 & 0.41 & 0.23 & 0.53 & 0.53 & 0.50 & 0.52 \\

Australian & 0.41 & 0.74 & 0.26 & 0.26 & 0.41 & 0.26 & 0.41 & 0.38 & 0.34 & 0.26 & 0.26 & 0.29 & 0.46 & \textbf{0.83} \\

LendingClub & 0.20 & 0.55 & 0.65 & 0.72 & 0.17 & 0.10 & 0.61 & 0.00 & 0.59 & 0.02 & 0.61 & 0.05 & 0.52 & \textbf{0.97} \\

ccf & 0.20 & 0.55 & 0.96 & 0.00 & 0.17 & 0.01 & 0.00 & \textbf{1.00} & \textbf{1.00} & 0.10 & 0.00 & 0.00 & 0.86 & \textbf{1.00} \\

ccfraud & 0.20 & 0.55 & 0.90 & 0.25 & 0.17 & 0.36 & 0.01 & 0.00 & 0.57 & 0.62 & 0.48 & 0.03 & 0.29 & \textbf{0.91} \\

Polish & 0.20 & 0.55 & 0.86 & \textbf{0.92} & 0.17 & 0.83 & \textbf{0.92} & 0.30 & \textbf{0.92} & 0.76 & \textbf{0.92} & 0.40 & 0.62 & \textbf{0.92} \\

Taiwan & 0.20 & 0.55 & \textbf{0.95} & \textbf{0.95} & 0.17 & 0.26 & \textbf{0.95} & 0.60 & 0.95 & 0.00 & 0.95 & 0.70 & 0.66 & \textbf{0.95} \\

Porto Seguro & 0.20 & 0.55 & 0.95 & 0.01 & 0.17 & 0.94 & 0.04 & \textbf{0.96} & \textbf{0.96} & 0.95 & 0.72 & 0.00 & 0.88 & 0.95 \\

Travel Insurance & 0.20 & 0.55 & 0.00 & 0.00 & 0.17 & 0.00 & 0.00 & \textbf{0.98} & 0.89 & 0.77 & 0.00 & 0.03 & 0.53 & \textbf{0.98} \\
\bottomrule
\end{tabularx}
\end{table*}

\section{Conclusion}
\label{sec:conclusion}
We proposed a novel RL-based approach to train LLMs for explainable tabular data prediction. Experimental results indicate the potential to improve prediction accuracy using a relatively small LLM on financial benchmark datasets, while also providing explanations for predictions. We view this as a preliminary step toward unlocking a wide range of research opportunities. Our framework is highly flexible and can be further enhanced by designing customized reward functions. For instance, other LLMs could be leveraged as judges or critics to provide feedback on the consistency and logical coherence of the responses. Additionally, evaluating the quality of generated explanations presents another exciting direction for future research. 

On the other hand, effectively addressing the challenges posed by highly imbalanced datasets during fine-tuning remains an area requiring further exploration. 
In parallel, another line of research is to improve the GRPO algorithm~\citep{Liu_drgrpo_2025, Yu_DAPO_2025}, which has shown potential in RL-based fine-tuning. Enhancing its stability, sample efficiency, and generalization capabilities—or developing entirely new reinforcement learning techniques specifically tailored for prediction tasks—could significantly advance the state of the art in the field.


\newpage
\bibliography{llm_tab}
\bibliographystyle{icml2025}

\newpage
\appendix
\onecolumn
\section{Examples of Generated Explanations and Answers}
\label{sec:generation-examples}

\begingroup
\small
\begin{longtable}{|l|l p{12cm}|}
\caption{Examples of generated reasoning and answer by TabReason}
\label{tab:generation-examples} \\
\hline
\textbf{Dataset} & \textbf{Example} &  \\
\hline
\endhead
German & Input Prompt & 
\vspace*{-\baselineskip} 
\begin{lstlisting}
Assess the creditworthiness of the following client as either 'good' or 'bad' based on the provided attributes.
The state of Status of existing checking account is bigger than 0 DM but smaller than 200 DM. The state of Duration in month is 48. The state of Credit history is all credits at this bank paid back duly. The state of Purpose is business. The state of Credit amount is 3566. The state of Savings account or bonds is bigger than 100 smaller than  500 DM. The state of Present employment since is bigger than 4  smaller than 7 years. The state of Installment rate in percentage of disposable income is 4. The state of Personal status and sex is male and single. The state of  Other debtors or guarantors is none. The state of Present residence since is 2. The state of Property is car or other. The state of Age in years is 30. The state of Other installment plans is none. The state of Housing is own. The state of Number of existing credits at this bank is 1. The state of Job is skilled employee or official. The state of Number of people being liable to provide maintenance for is 1. The state of Telephone is none. The state of foreign worker is yes.    
\end{lstlisting}
\\
       & LLM Output & 
\vspace*{-\baselineskip} 
\begin{lstlisting}
<reasoning>
Based on the provided attributes, the client meets several criteria for good creditworthiness:
1. The duration in months is very long (48 months), indicating a stable financial situation over time.
2. The purpose of the loan is for business, which is typically considered a less risky decision for credit purposes.
3. The credit amount is relatively high (3566 DM), but within a moderate range.
4. The client has a good credit history with all payments made on time.
5. The client is employed for an extended period (4+ years).
6. There are few liabilities: the client owns their property and has a skilled employment status.
7. The client has no other debts.
This combination of factors suggests strong financial stability and a history of reliable payment. The personal situation (male, single, foreign worker) aligns well with most clients' profiles.
</reasoning>

<answer>good</answer>
\end{lstlisting}
\\
\hline
LendingClub & Input Prompt & 
\vspace*{-\baselineskip} 
\begin{lstlisting}
Assess the client's loan status as either 'good' or 'bad' based on the following loan records from Lending Club.
The client has attributes as follows: The state of Installment is 765.37. The state of Loan Purpose is home_improvement. The state of Loan Application Type is Individual. The state of Interest Rate is 22.15%. The state of Last Payment Amount is 765.37. The state of Loan Amount is 20000.0. The state of Revolving Balance is 24472.0. The state of Delinquency In 2 years is 1.0. The state of Inquiries In 6 Months is 2.0. The state of Mortgage Accounts is 2.0. The state of Grade is E. The state of Open Accounts is 26.0. The state of Revolving Utilization Rate is 82.1%. The state of Total Accounts is 34.0. The state of Fico Range Low is 660.0. The state of Fico Range High is 664.0. The state of Address State is IL. The state of Employment Length is 6 years. The state of Home Ownership is MORTGAGE. The state of Verification Status is Verified. The state of Annual Income is 57000.0.
\end{lstlisting}
\\
            & LLM Output & 
\vspace*{-\baselineskip} 
\begin{lstlisting}
<reasoning>
Based on the given loan records, we can assess the client's loan status as 'bad'. Here's the reasoning:

- The client's grade (E) indicates a high risk category for subprime loans.
- The interest rate of 22.15% suggests higher-than-average borrowing costs compared to more stable clients.
- The delinquency in two years (1.0) shows that the client has had some payment issues but not severe enough to be considered bad.
- However, multiple inquiries within six months (2.0) could indicate repeated applications or attempts to refinance, which might suggest underlying creditworthiness concerns.

Given these factors, especially the grade and the delinquency rate, it seems prudent to classify this loan application under the 'bad' category due to the elevated risk profile indicated by its low grade despite having minimal delinquencies.
</reasoning>

<answer>Bad</answer>
\end{lstlisting}
\\
\hline
Travel Insurance & Input Prompt & 
\vspace*{-\baselineskip} 
\begin{lstlisting}
Determine the claim status as either `yes' or `no' based on the following table attributes for travel insurance status. The table attributes including 5 categorical attributes and 4 numerical attributes are as follows:
Agency: Name of agency (categorical).
Agency Type: Type of travel insurance agencies (categorical).
Distribution Channel: Distribution channel of travel insurance agencies (categorical).
Product Name: Name of the travel insurance products (categorical).
Duration: Duration of travel (numerical).
Destination: Destination of travel (categorical).
Net Sales: Amount of sales of travel insurance policies (numerical).
Commission: Commission received for travel insurance agency (numerical).
Age: Age of insured (numerical).
The insurance company has attributes: Agency: JZI, Agency Type: Airlines, Distribution Channel: Online, Product Name: Value Plan, Duration: 88, Destination: NETHERLANDS, Net Sales: 125.0, Commission: 43.75, Age: 24.
\end{lstlisting}
\\
            & LLM Output & 
\vspace*{-\baselineskip} 
\begin{lstlisting}
<reasoning>
The given table contains various attributes related to a specific travel insurance product from JZI's online distribution channel. Let's analyze each attribute:

- **Agency**: JZI - This indicates that the insurance company is JZI.
- **Agency Type**: Airlines - Indicates it is an airline-related policy.
- **Distribution Channel**: Online - Specifies the way the policy was sold.
- **Product Name**: Value Plan - Identifies the name of the insurance product.
- **Duration**: 88 - Indicates the duration of the trip.
- **Destination**: NETHERLANDS - Indicates where the destination is located.
- **Net Sales**: 125.0 - Represents the amount of money earned through this policy.
- **Commission**: 43.75 - Represents the commission earned per unit of sale.

Based on these details, we can conclude that JZI offers a value plan insurance product with a duration of 88 days to a Netherlands tourist. However, there isn't enough information about their age limit or other specific conditions of the policy beyond what's provided.

Given the limited data available, it's difficult to make a definitive judgment on whether this claim statement accurately reflects the current status of the travel insurance policy without more context or additional information such as coverage limits, exclusions, etc.

Therefore, the answer would be ``No'' because while some basic facts like the insurer, type of policy, and terms can be inferred, making a complete assessment requires deeper analysis which isn't possible solely from the provided information.
</reasoning>

<answer>No</answer>
\end{lstlisting}
\\
\hline
Taiwan & Input Prompt & 
\vspace*{-\baselineskip} 
\begin{lstlisting}[mathescape=true]
Predict whether the company will face bankruptcy as either 'yes' or 'no' based on the following financial attributes.     
The client has attributes: Bankrupt?: 0.409,  ROA(C) before interest and depreciation before interest: 0.420,  ROA(A) before interest and % after tax: 0.465,  ROA(B) before interest and depreciation after tax: 0.598,  Operating Gross Margin: 0.598,  Realized Sales Gross Margin: 0.999,  Operating Profit Rate: 0.797,  Pre-tax net Interest Rate: 0.809,  After-tax net Interest Rate: 0.303,  Non-industry income and expenditure/revenue: 0.781,  Continuous interest rate (after tax): 0.000,  Operating Expense Rate: 9290000000.000,  Research and development expense rate: 0.453,  Cash flow rate: 0.000,  Interest-bearing debt interest rate: 0.000,  Tax rate (A): 0.169,  Net Value Per Share (B): 0.169,  Net Value Per Share (A): 0.169,  Net Value Per Share (C): 0.193,  Persistent EPS in the Last Four Seasons: 0.301,  Cash Flow Per Share: 0.029,  Revenue Per Share (Yuan $\yen$): 0.090,  Operating Profit Per Share (Yuan $\yen$): 0.134,  Per Share Net profit before tax (Yuan $\yen$): 0.022,  Realized Sales Gross Profit Growth Rate: 0.848,  Operating Profit Growth Rate: 0.687,  After-tax Net Profit Growth Rate: 0.687,  Regular Net Profit Growth Rate: 0.217,  Continuous Net Profit Growth Rate: 5610000000.000,  Total Asset Growth Rate: 0.000,  Net Value Growth Rate: 0.262,  Total Asset Return Growth Rate Ratio: 0.337,  Cash Reinvestment %: 0.007,  Current Ratio: 0.007,  Quick Ratio: 0.631,  Interest Expense Ratio: 0.009,  Total debt/Total net worth: 0.145,  Debt ratio %: 0.855,  Net worth/Assets: 0.007,  Long-term fund suitability ratio (A): 0.376,  Borrowing dependency: 0.008,  Contingent liabilities/Net worth: 0.090,  Operating profit/Paid-in capital: 0.133,  Net profit before tax/Paid-in capital: 0.407,  Inventory and accounts receivable/Net value: 0.133,  Total Asset Turnover: 0.000,  Accounts Receivable Turnover: 0.014,  Average Collection Days: 0.001,  Inventory Turnover Rate (times): 0.001,  Fixed Assets Turnover Frequency: 0.034,  Net Worth Turnover Rate (times): 0.038,  Revenue per person: 0.387,  Operating profit per person: 0.004,  Allocation rate per person: 0.770,  Working Capital to Total Assets: 0.560,  Quick Assets/Total Assets: 0.551,  Current Assets/Total Assets: 0.036,  Cash/Total Assets: 0.007,  Quick Assets/Current Liability: 0.001,  Cash/Current Liability: 0.149,  Current Liability to Assets: 0.317,  Operating Funds to Liability: 0.277,  Inventory/Working Capital: 0.000,  Inventory/Current Liability: 0.975,  Current Liabilities/Liability: 0.734,  Working Capital/Equity: 0.334,  Current Liabilities/Equity: 66300000.000,  Long-term Liability to Current Assets: 0.917,  Retained Earnings to Total Assets: 0.002,  Total income/Total expense: 0.059,  Total expense/Assets: 0.000,  Current Asset Turnover Rate: 0.000,  Quick Asset Turnover Rate: 0.594,  Working capitcal Turnover Rate: 2770000000.000,  Cash Turnover Rate: 0.672,  Cash Flow to Sales: 0.068,  Fixed Assets to Assets: 0.975,  Current Liability to Liability: 0.334,  Current Liability to Equity: 0.111,  Equity to Long-term Liability: 0.633,  Cash Flow to Total Assets: 0.458,  Cash Flow to Liability: 0.455,  CFO to Assets: 0.313,  Cash Flow to Equity: 0.042,  Current Liability to Current Assets: 0.000,  Liability-Assets Flag: 0.726,  Net Income to Total Assets: 0.002,  Total assets to GNP price: 0.624,  No-credit Interval: 0.598,  Gross Profit to Sales: 0.834,  Net Income to Stockholder's Equity: 0.281,  Liability to Equity: 0.027,  Degree of Financial Leverage (DFL): 0.565,  Interest Coverage Ratio (Interest expense to EBIT): 1.000,  Net Income Flag: 0.025. 
\end{lstlisting}
\\
       & LLM Output & 
\vspace*{-\baselineskip} 
\begin{lstlisting}
<reasoning>
Based on the financial attributes provided, the company appears to have a strong financial position. The company's key financial ratios are all within healthy ranges, indicating positive cash flow, low debt levels, and efficient operations. Additionally, the company has a positive operating profit growth rate ("687%") which suggests that the business is growing and is able to maintain profitability. The company's debt-to-equity ratio is also below 50%, indicating a relatively low level of financial risk. 
While the company has a very high after-tax net interest rate, which could be indicative of a company with significant long-term debt, this is not enough to suggest bankruptcy in and of itself. The company also has a high "Net Income Flag" of 0.025, indicating that the company is able to generate a positive net income. 
Overall, based on the strong performance of the company in terms of financial ratios and the ability to maintain profitability, it is unlikely that the company will face bankruptcy. However, it is always important to assess all financial attributes on an ongoing basis and to remain vigilant for any changes in the company's financial status.
</reasoning>
<answer>
no
</answer>
\end{lstlisting}
\\
\hline
\end{longtable}
\endgroup

\section{RL Fine-Tuning}
\label{sec:rl-tuning}
For RL fine-tuning, the model was trained for up to 50 epochs or a maximum duration of 60 hours on a single A100 GPU. During training, the LLM parameters were set to \texttt{temperature = 0.7}, \texttt{top-p = 0.8}, and \texttt{top\text-k = 20}. However, during inference, we found that using a lower temperature of \texttt{0.1} was more effective, as it shifted the learned policy toward greater exploitation rather than exploration. For the results in Table~\ref{tab:llm-performance}, the best epoch was selected based on the best weighted F1 score on validation set.

\begin{figure}[htb]
     \centering
     \begin{subfigure}[t]{0.45\textwidth}
         \centering
         \includegraphics[width=\textwidth]{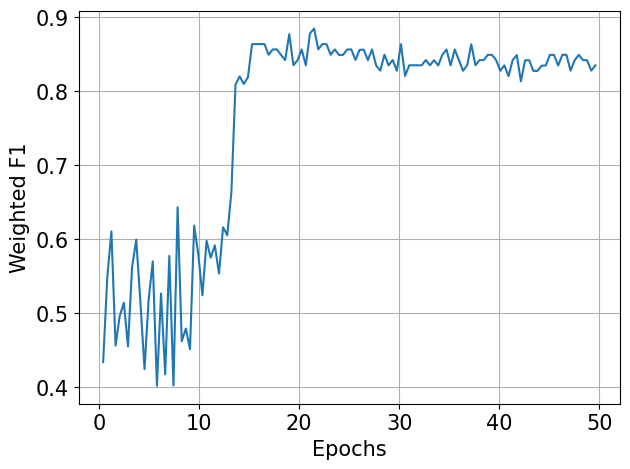}
         \caption{Australian}
     \end{subfigure}
     \hfill
     \begin{subfigure}[t]{0.45\textwidth}
         \centering
         \includegraphics[width=\textwidth]{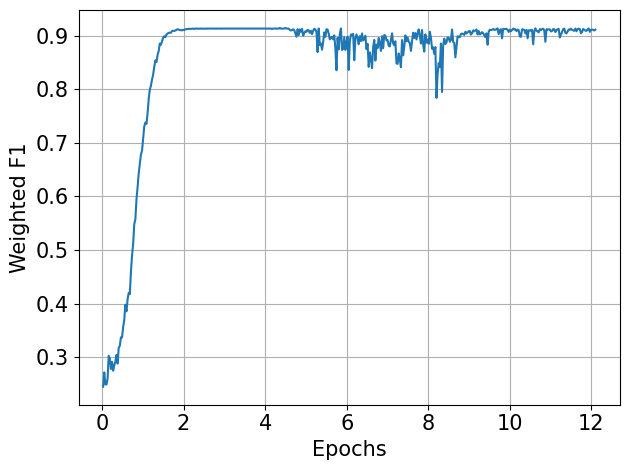}
         \caption{ccfraud}
     \end{subfigure}
     \hfill
     \begin{subfigure}[t]{0.45\textwidth}
         \centering
         \includegraphics[width=\textwidth]{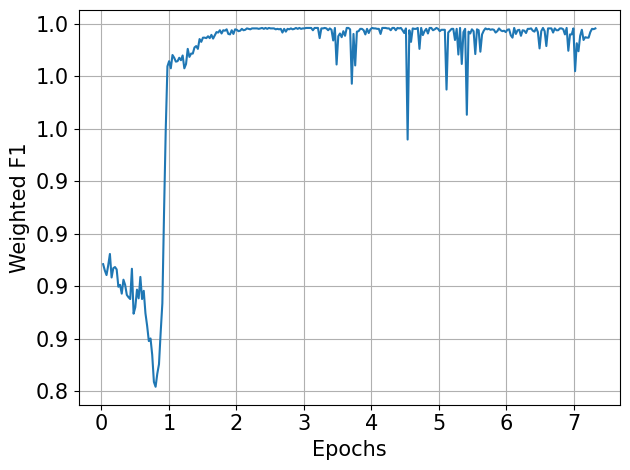}
         \caption{ccf}
     \end{subfigure}
     \hfill
     \begin{subfigure}[t]{0.45\textwidth}
         \centering
         \includegraphics[width=\textwidth]{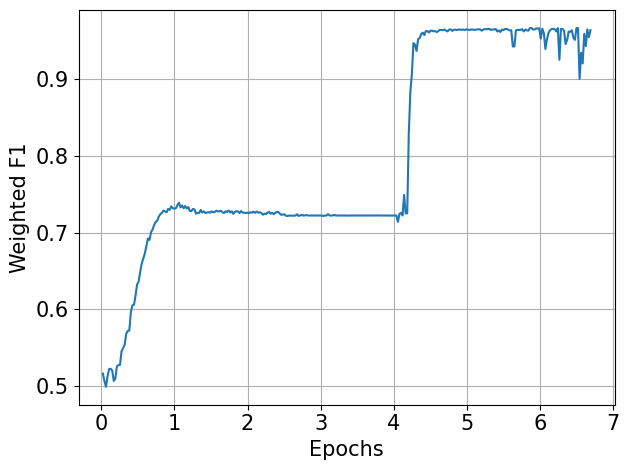}
         \caption{LendingClub}
     \end{subfigure}
        \caption{Examples of model performance over epochs using the proposed RL fine-tuning model.}
        \label{fig:weighted_f1_over_epochs}
\end{figure}

\section{Pitfalls}
\label{sec:challenges}
We found two main pitfalls in the experiments describe in Section~\ref{sec:experiments}:
\begin{itemize}
\item Imbalanced Labels: 
For datasets such as CCF, CCFraud, Polish, and Travel Insurance, where the labels are highly imbalanced, RL fine-tuning increases overall accuracy and weighted F1 score but tends to converge on predicting the majority class. We attempted to address this by applying inversely weighted rewards to balance the model, but this approach was not successful. In contrast, RL fine-tuning performed very well on the LendingClub and Australian datasets, where the label distribution is more balanced.

To better demonstrate TabReason’s performance on imbalanced datasets, we also evaluate results using the Matthews Correlation Coefficient (MCC), which provides a balanced assessment of binary classification quality by considering true and false positives and negatives, even in the presence of class imbalance. The results for both the weighted F1 score and MCC are presented in Table~\ref{tab:llm-performance-f1-mcc}.

\begin{table*}[!htb]
\caption{Performance comparison of various LLMs across different financial datasets using weighted F1 score and MCC metric.}
\label{tab:llm-performance-f1-mcc}
\centering
\scriptsize
\setlength{\tabcolsep}{2.2pt}
\begin{tabularx}{1.02\textwidth}{L{46pt} L{17pt} C{20pt} C{20pt} *{9}{C{26pt}} C{35pt} C{26pt} C{24pt}}
\toprule
\textbf{Dataset} & \textbf{Metric} & \textbf{Chat-GPT} & \textbf{GPT-4} & \textbf{Gemini} & \textbf{Llama2-7B-chat} & \textbf{Llama2-70B} & \textbf{Llama3-8B} & \textbf{FinMA-7B} & \textbf{FinGPT-7B-lora} & \textbf{InternLM-7B} & \textbf{Falcon-7B} & \textbf{Mixtral-7B} & \textbf{CFGPT-sft-7B-Full} & \textbf{Qwen2.5-1.5B} & \textbf{TabReason} \\
\midrule
German & F1 & 0.20 & 0.55 & 0.52 & \textbf{0.57} & 0.17 & 0.56 & 0.17 & 0.52 & 0.41 & 0.23 & 0.53 & 0.53 & 0.50 & 0.52 \\
       & MCC & -0.10 & -0.02 & 0.00 & 0.03 & 0.00 & \textbf{0.05} & 0.00 & 0.00 & -0.30 & -0.07 & 0.00 & 0.00 & -0.12 & -0.06 \\

Australian & F1 & 0.41 & 0.74 & 0.26 & 0.26 & 0.41 & 0.26 & 0.41 & 0.38 & 0.34 & 0.26 & 0.26 & 0.29 & 0.46 & \textbf{0.83} \\
           & MCC & 0.00 & \textbf{0.47} & 0.00 & 0.00 & 0.00 & 0.00 & 0.00 & 0.11 & 0.13 & 0.00 & 0.00 & -0.10 & -0.05 & \textbf{0.66} \\

LendingClub & F1 & 0.20 & 0.55 & 0.65 & 0.72 & 0.17 & 0.10 & 0.61 & 0.00 & 0.59 & 0.02 & 0.61 & 0.05 & 0.52 & \textbf{0.97} \\
            & MCC & -0.10 & -0.02 & \textbf{0.19} & 0.00 & 0.00 & -0.15 & 0.00 & 0.00 & 0.15 & -0.01 & 0.08 & 0.01 & 0.05 & \textbf{0.89} \\

ccf & F1 & 0.20 & 0.55 & 0.96 & 0.00 & 0.17 & 0.01 & 0.00 & \textbf{1.00} & \textbf{1.00} & 0.10 & 0.00 & 0.00 & 0.86 & \textbf{1.00} \\
     & MCC & -0.10 & -0.02 & -0.01 & \textbf{0.00} & 0.00 & 0.00 & 0.00 & 0.00 & 0.00 & 0.00 & 0.00 & 0.00 & -0.02 & 0.00 \\

ccfraud & F1 & 0.20 & 0.55 & 0.90 & 0.25 & 0.17 & 0.36 & 0.01 & 0.00 & 0.57 & 0.62 & 0.48 & 0.03 & 0.29 & \textbf{0.91} \\
        & MCC & -0.10 & -0.02 & 0.00 & -0.16 & 0.00 & -0.03 & -0.06 & 0.00 & -0.13 & -0.02 & \textbf{0.16} & 0.01 & 0.02 & 0.00 \\

Polish & F1 & 0.20 & 0.55 & 0.86 & \textbf{0.92} & 0.17 & 0.83 & \textbf{0.92} & 0.30 & \textbf{0.92} & 0.76 & \textbf{0.92} & 0.40 & 0.62 & \textbf{0.92} \\
       & MCC & -0.10 & -0.02 & \textbf{0.14} & 0.00 & 0.00 & -0.06 & -0.01 & 0.00 & 0.07 & 0.05 & 0.00 & -0.02 & 0.05 & 0.04 \\

Taiwan & F1 & 0.20 & 0.55 & \textbf{0.95} & \textbf{0.95} & 0.17 & 0.26 & \textbf{0.95} & 0.60 & 0.95 & 0.00 & 0.95 & 0.70 & 0.66 & \textbf{0.95} \\
       & MCC & -0.10 & -0.02 & \textbf{0.00} & -0.01 & 0.00 & -0.07 & 0.00 & -0.02 & -0.01 & 0.00 & 0.00 & 0.00 & -0.05 & 0.00 \\

Porto Seguro & F1 & 0.20 & 0.55 & 0.95 & 0.01 & 0.17 & 0.94 & 0.04 & \textbf{0.96} & \textbf{0.96} & 0.95 & 0.72 & 0.00 & 0.88 & 0.95 \\
            & MCC & -0.10 & -0.02 & 0.00 & -0.05 & 0.00 & -0.01 & \textbf{0.01} & 0.00 & 0.00 & 0.00 & 0.01 & 0.00 & -0.02 & 0.00 \\

Travel Insurance & F1 & 0.20 & 0.55 & 0.00 & 0.00 & 0.17 & 0.00 & 0.00 & \textbf{0.98} & 0.89 & 0.77 & 0.00 & 0.03 & 0.53 & \textbf{0.98} \\
                & MCC & -0.10 & -0.02 & 0.00 & 0.00 & 0.00 & 0.00  & 0.00 & 0.00 & \textbf{0.12} & -0.03 & 0.00 & 0.01 & 0.03 & 0.00 \\
\bottomrule
\end{tabularx}
\end{table*}

\item Inconsistency between reasoning and final answer: 
We observed some instances of inconsistencies between the reasoning component and the final answer. This may be attributed to the small size of the Qwen model we use, as well as the use of a non-zero temperature. However, our experiments showed that setting the temperature to zero reduces both the quality of generated responses and prediction performance, so this is not a viable solution for resolving inconsistencie
\end{itemize}


\end{document}